\documentclass[conference]{IEEEtran}
\IEEEoverridecommandlockouts
\usepackage{cite}
\usepackage{amsmath,amsfonts}
\usepackage{bbm}
\usepackage{graphicx}
\usepackage{textcomp}
\usepackage{lipsum}
\usepackage{xcolor}
\usepackage{mathtools}
\usepackage{float}
\usepackage{algorithm}
\usepackage{algpseudocode}
\usepackage{amssymb}

\usepackage{array}
\newcolumntype{d}[1]{>{\centering\arraybackslash}m{#1\linewidth}}

\DeclareMathOperator*{\argmax}{argmax}
\def\BibTeX{{\rm B\kern-.05em{\sc i\kern-.025em b}\kern-.08em
T\kern-.1667em\lower.7ex\hbox{E}\kern-.125emX}}
\begin{document}

    \title{\huge Reliability-Optimized User Admission Control for URLLC Traffic: A Neural Contextual Bandit Approach
\thanks{© 2024 IEEE. Personal use of this material is permitted. Permission from IEEE must be
	obtained for all other uses, in any current or future media, including
	reprinting/republishing this material for advertising or promotional purposes, creating new
	collective works, for resale or redistribution to servers or lists, or reuse of any copyrighted
	component of this work in other works.}
    }

    \author{\IEEEauthorblockN{
        Omid Semiari%\IEEEauthorrefmark{1}
        , Hosein Nikopour
        %\IEEEauthorrefmark{2}
        and Shilpa Talwar
    %\IEEEauthorrefmark{3}
    }\IEEEauthorblockA{%\IEEEauthorrefmark{1}
        Intel Labs, Santa Clara, CA\\Email: \{omid.semiari, hosein.nikopour, shilpa.talwar\}@intel.com\vspace{-1em}}
    %\IEEEauthorblockA{\IEEEauthorrefmark{2}Twentieth Century Fox, Springfield, USA\\Email: homer@thesimpsons.com}
    }

    \maketitle
    
    \thispagestyle{plain} % for the first page
    \pagestyle{plain}     % for the rest of the pages

    \begin{abstract}
         Ultra-reliable low-latency communication (URLLC) is the cornerstone for a broad range of emerging services in next-generation wireless networks. URLLC fundamentally relies on the network's ability to proactively determine whether sufficient resources are available to support the URLLC traffic, and thus, prevent so-called cell overloads. Nonetheless, achieving accurate quality-of-service (QoS) predictions for URLLC user equipment (UEs) and preventing cell overloads are very challenging tasks. This is due to dependency of the QoS metrics (latency and reliability) on traffic and channel statistics, users' mobility, and interdependent performance across UEs. In this paper, a new \emph{QoS-aware UE admission control} approach is developed to proactively estimate QoS for URLLC UEs, prior to associating them with a cell, and accordingly, admit only a subset of UEs that do not lead to a cell overload. To this end, an optimization problem is formulated to find an efficient UE admission control policy, cognizant of UEs' QoS requirements and cell-level load dynamics. To solve this problem, a new machine learning based method is proposed that builds on (deep) neural contextual bandits, a suitable framework for dealing with nonlinear bandit problems. In fact, the UE admission controller is treated as a bandit agent that observes a set of network measurements (context) and makes admission control decisions based on context-dependent QoS (reward) predictions.  The simulation results show that the proposed scheme can achieve near-optimal performance and yield substantial gains in terms of cell-level service reliability and efficient resource utilization.
    \end{abstract}

    %\begin{IEEEkeywords}
    %    component, formatting, style, styling, insert
    %\end{IEEEkeywords}

    \section{Introduction}~\label{sec:intro}
    %\lipsum[1-2]
    Ultra-reliable low-latency communication (URLLC) is anticipated to be an integral part of the sixth generation (6G) wireless networks, enabling new wireless edge services such as cloud-based augmented reality and gaming, remote control, vehicle-to-infrastructure communications (V2X), mission critical Internet of things (IoT) and real-time digital twins~\cite{DBLP, 8869705, 8642794}.  For these services, the network must reliably meet extreme latency constraints of user and control plane transmissions.

     In URLLC, an imperative open problem is to \emph{accurately} predict quality-of-service (QoS), in terms of reliability and latency metrics, before admitting new user equipment (UEs) to a cell. Such QoS predictions can be used by the cell to ensure enough resources are available to meet strict latency and reliability requirements of URLLC UEs and proactively eliminate so-called “cell overloads”. Achieving accurate QoS and cell overload predictions for URLLC is a  challenging problem in realistic complex network scenarios. This is due to the fact that reliability is intrinsically linked to the statistical distribution of the transmission delay~\cite{lopez2022statistical, 8673783}. However, delay distributions are generally unknown to the cell and impacted by various sources of uncertainties in wireless networks such as channel fading, interference, noise, non-deterministic traffic, and users’ mobility. Hence, the cell must \emph{learn} a reasonably accurate QoS model that can capture those network dynamics.

    For typical voice and data traffic, cell overloads are traditionally managed by the call admission control  (CAC)~\cite{7556057} mechanism to limit the number of UEs and QoS flows served by a cell.  In~\cite{1509946}, the authors provide an overview of existing CAC methods. Despite being widely adopted, these dynamic CAC schemes rely on simple rule-based policies, e.g., based on bandwidth utilization thresholds~\cite{5396157}, service prioritization~\cite{5426190}, or call dropping rate~\cite{840209}. In fact, none of the methods in~\cite{1509946, 5396157, 5426190, 840209} directly addresses the admission control for URLLC traffic. Complementary to the CAC, network slice admission control (NSAC) is introduced by 3GPP~\cite{3gpp:TS29.536} which heuristically controls the number of UEs and packet data unit (PDU) sessions registered to a network slice. The works in~\cite{9129141} and~\cite{10195485} develop admission control schemes for single-cell scenarios with both URLLC and mobile broadband UEs. In~\cite{9129141}, the URLLC traffic is prioritized by admitting all URLLC UEs and only applying the admission control to mobile broadband UEs. That is, the  method in~\cite{9129141} does not directly address the admission control for URLLC. Moreover, the delay and reliability models presented in~\cite{9129141} and~\cite{10195485}   rely on overly simplified assumptions (e.g., static channel, no packet re-transmissions) that do not hold in real mobile networks.
    
    Regarding QoS modeling and prediction for URLLC, the authors in~\cite{lopez2022statistical} and~\cite{8673783} provide extensive statistical analyses of latency and reliability metrics by using an array of theoretical tools such as parameterized channel models, tail approximations, and risk assessment concepts. Despite their fundamental results, derived QoS analysis mostly apply to simple cases and do not easily extend to more complex network scenarios with radio link control (RLC) packet re-transmissions, users' mobility, and multi-user scheduling. Moreover, these works do not focus on the admission control for URLLC traffic. In addition to model-based approaches, there is a recent body of work in~\cite{9685108, 9625281, 9348128} that develops machine learning (ML) based solutions for directly deriving an admission policy. The works in ~\cite{9685108} and~\cite{9625281} leverage  different deep reinforcement learning techniques to learn admission policies for non-URLLC traffic. Moreover,~\cite{9348128}  uses Q-learning to address  admission control in a generic service system with delay constraints. However, the adopted model does not consider real radio access networks (RANs) constraints associated with the fading channel, packet transmission failures, and bandwidth limitations.

    The main contribution of this paper is a novel ML-based solution to address the \emph{QoS-aware user admission control} for URLLC traffic in open RAN (O-RAN) networks.  The proposed solution builds on (deep) neural contextual bandits~\cite{Allesiardo_NeuralCMAB}, a powerful ML framework that deals with complex decision-making problems under uncertainties. In particular, the proposed solution treats the admission controller as a contextual bandit (CB) agent that can be implemented as an xApp in O-RAN. Upon receiving service requests from applicant UEs (either handover or activated URLLC UEs), the xApp observes RAN measurements (contextual information) associated with UEs and the O-RAN radio unit (O-RU), and accordingly, admits a subset of applicants. To optimize the admission control policy, the proposed method employs a system of deep neural networks (DNNs) to predict the service \emph{reliability as a nonlinear function of the context} for each admission decision. A key advantage of the proposed approach is to handle concurrent admission requests from multiple UEs. Further, unlike existing model-based admission control methods that depend on strong simplifying assumptions about the network, the agent in our proposed approach directly learns from experience (context-reward observations). As shown by comprehensive simulations, the learned admission control policy for URLLC traffic yields near-optimal performance and generalizes well to complex O-RAN scenarios with practical system-level considerations such as mini-slot configurations, packet re-transmissions, UE mobility, and link adaptation with imperfect channel information.

    The rest of the paper is organized as follows. Sec.~\ref{sec:system_model}  presents the system model and problem formulation. Sec.~\ref{sec:MAB} formally defines the UE admission control as a neural CB and presents the proposed solution. Simulation results are included in Sec.~\ref{sec:simulations} and
    conclusions are outlined in Sec.~\ref{sec:conclusions}.

    \section{System Model}\label{sec:system_model}
\begin{figure}
	\centering
	\centerline{\includegraphics[width=7cm]{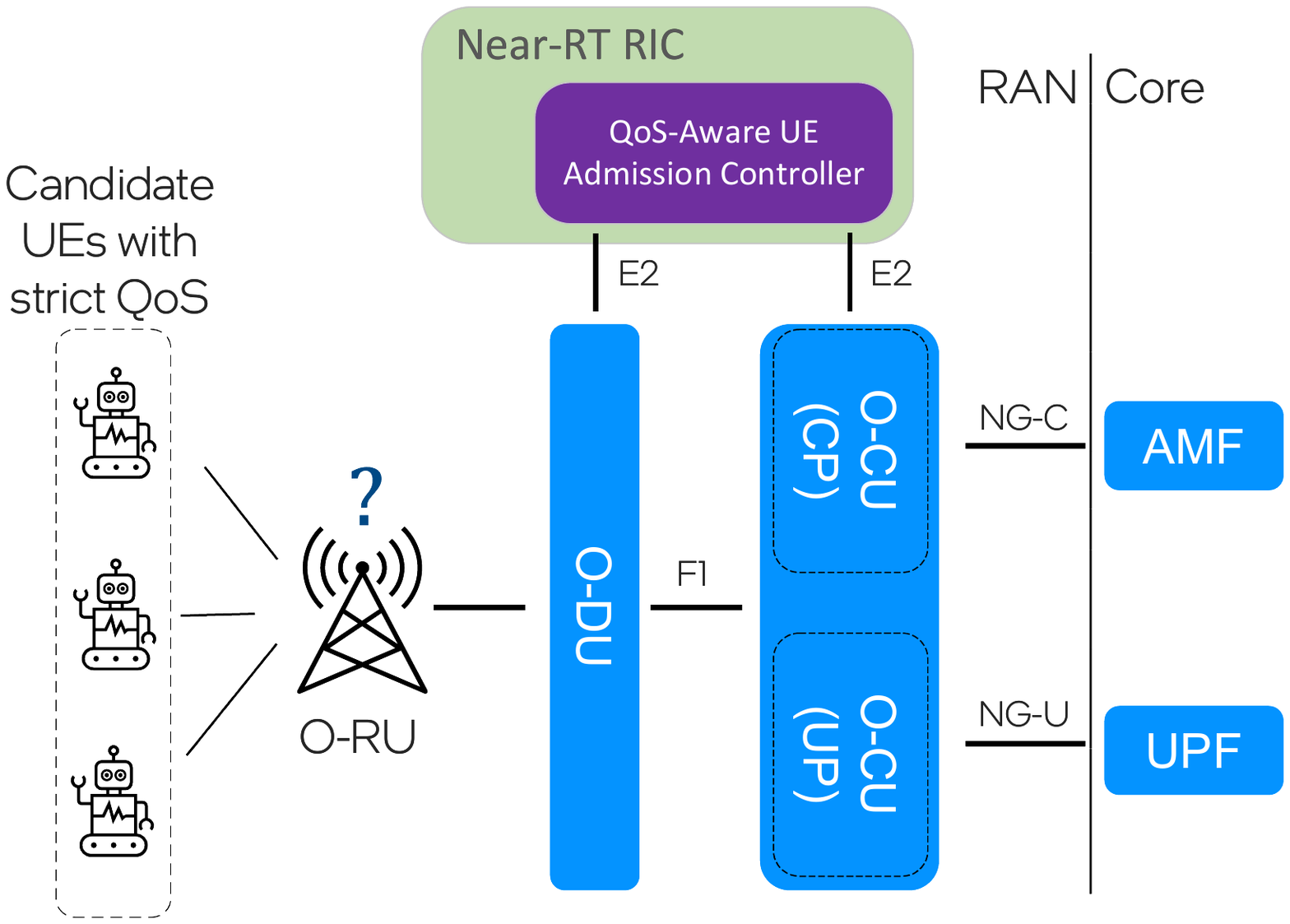}}\vspace{-1em}
	\caption{QoS-aware UE admission controller as an xApp in O-RAN.}
	\label{model}\vspace{-0.5em}
\end{figure}

   Consider $K$ URLLC UEs, in a set $\mathcal{K}$, randomly located within the coverage of an O-RU cell.  As shown in Fig.~\ref{model}, the cell may receive new service requests from UEs under two conditions: (1) UE is already associated with the cell but it is transitioning from  one of radio resource control (RRC) ``inactive'' or RRC ``idle'' modes to an RRC ``connected'' mode, and (2) UE is requesting to perform a handover to the cell (as a target cell).  In either case, before admitting new service requests from UEs, the cell must employ a QoS-aware admission control to proactively determine whether enough resources are available to meet the QoS for all admitted UEs, given the current cell load and other RAN measurements. If there are not enough resources available, then the cell must only admit a subset of UEs for whom the QoS can be met. 
   
   \subsection{QoS Metrics, Scheduling, and Traffic Model for URLLC}
   
    Let $\mathcal{K}' \!\subseteq\! \mathcal{K}$ and $\mathcal{K}'' \!\subseteq\! \mathcal{K}$ denote, respectively, the set of UEs submitting new service requests, hereinafter called \emph{applicant UEs}, and the set of currently active UEs  in the cell (i.e., UEs in the RRC connected mode). The QoS for an arbitrary URLLC UE $k \in \mathcal{K}$ can be defined as:
   \begin{align}\label{qos_per_ue}
    \mathbb{P}(t_k \leq \tau_k) \geq \delta_k,
   \end{align}
where $t_k$ represents the transmission delay over the access link. Aligned with the 3rd generation partnership project (3GPP) definition for user plane latency~\cite{3gpp:TR38.913}, the transmission delay is defined here as the radio interface latency from the time when the cell's packet data convergence protocol (PDCP) receives an Internet Protocol (IP) packet to the time when UE successfully receives the IP packet. Moreover, $\delta_k \in [0,1]$ and $\tau_k$ represent, respectively, the reliability and delay requirements; both are determined based on the specific use case and URLLC QoS requirements. Without loss of generality, let’s assume $\delta_k=\delta$ for all $k=1, \cdots, K$.

Moreover, we consider a non-full buffer traffic by modeling packet arrival for each UE $k \in \mathcal{K}' \cup \mathcal{K}''$ as a Poisson process with a packet arrival rate of  $\lambda_{k}$ and a PDCP packet size of $b_k$. The cell employs link adaptation to adjust the rate according to non-ideal periodic channel state information (CSI) feedback received from UEs. 
%\textcolor{blue}{We note that packet retransmissions at the PDCP layer are generally avoided in URLLC due to introducing excessive delays;} 
We consider retransmissions at the RLC layer. That is, if the transmission of a transport block is failed, transmitted bits are scheduled for retrainsmission during next transmission time intervals (TTIs), unless the delay $t_k$ exceeds the tolerable delay $\tau_k$. In that case, the packet is dropped from the queue. For allocating resource block groups (RBGs) to admitted UEs, we consider maximum largest weighted delay first (M-LWDF) scheduler which is known to maintain a good balance
between QoS provisioning and fairness. At each TTI, the M-LWDF scheduler assigns RBGs (one-by-one) to any UE with a non-empty buffer that maximizes the following metric:
\begin{align}\label{m-lwdf}
	m_k = \zeta_k d_{\text{HOL}, k}\frac{c_k(t)}{\bar{c}_k(t-1)}.
\end{align}
In \eqref{m-lwdf}, $\zeta_k = -\log(1-\delta)/\tau_k$ and $d_{\text{HOL}, k}$ represents the delay of ``head-of-the-line (HOL)'', i.e., UE $k$'s oldest packet in the buffer. Moreover, $c_k(t)$ and $\bar{c}_k(t-1)$ denote, respectively, the achievable data rate at TTI $t$ and the average rate for UE $k$.  

\subsection{QoS-Aware User Admission Control Policy}
As shown in Fig.~\ref{model}, the user admission controller can be implemented as an xApp at the near real-time RAN intelligent controller (near-RT RIC)  in O-RAN. In particular, upon receiving new service requests from UEs, a cell can forward a set of relevant RAN data, hereinafter called the \emph{network context}, to the xApp to make an admission control decision. The network context (to be concretely defined in Sec.~\ref{Sec:III-B}) includes a set of UE-level and cell-level RAN measurements that are relevant to making QoS-aware admission control decisions. With this in mind, we can define the QoS-aware UE admission control as a policy $\pi:\mathbb{R}^{|\boldsymbol{x}|}\rightarrow \{0, 1\}^{|\mathcal{K}'|}$ that maps the network context vector $\boldsymbol{x}$ to an admission decision vector $\boldsymbol{z}\in \{0, 1\}^{|\mathcal{K}'|}$ with elements $z_{k'}$, $k'\in \mathcal{K}'$. Here, $z_{k'}=1$ if the service request of UE $k'$ is accepted, otherwise $z_{k'}=0$. Using this model, we can define the UE’s reliability indicator function as follows:
\begin{align}\label{ue_reliability}
	\mathbbm{1}(k, \boldsymbol{x}, \boldsymbol{z}) = \begin{cases}
		1, & \text{if}\, F_{t_k}(\tau_k; \boldsymbol{x}, \boldsymbol{z}) \geq \delta,\\
		0, & \text{otherwise},
	\end{cases} 
\end{align}
where $F_{t_k}(\tau_k; \boldsymbol{x}, \boldsymbol{z}) =  \mathbb{P}(t_k \leq \tau_k | \boldsymbol{x}, \boldsymbol{z})$ is the conditional cumulative distribution function (CDF) of $t_k$ at $\tau_k$. Next, we propose an optimization problem that builds on \eqref{ue_reliability} to maximize the service reliability at the cell level.

\subsection{Problem Formulation}
To optimize the admission control for URLLC UEs, let the policy $\pi$ be parameterized by a set of weights $\boldsymbol{\theta}$, thus, $\boldsymbol{z}=\pi(\boldsymbol{x}; \boldsymbol{\theta})$. Considering $\mathcal{K}_a=\left\{k'\in \mathcal{K}' | z_{k'}=1\right\}$ as the set of  accepted applicant UEs, we can define the cell reliability as
%\begin{align}\label{cell_reliability}
%	\!\!R(\boldsymbol{x}, \boldsymbol{\theta}) \!=\! \frac{K_a}{K}\!\prod_{k \in \mathcal{K}_a}{\!\!\!\mathbbm{1}(k, \boldsymbol{x}, \boldsymbol{z}) } \!=\! \frac{K_a}{K}\overbrace{\prod_{k \in \mathcal{K}_a \cup \mathcal{K}''}{\!\!\mathbbm{1}(k, \boldsymbol{x}, \pi(\boldsymbol{x}; \boldsymbol{\theta})), }}^{\text{cell reliability indicator}}
%\end{align}
\begin{align}\label{cell_reliability}
	\mathbbm{1}_c(\boldsymbol{x}, \boldsymbol{\theta}) =\!\!\! \prod_{k \in \mathcal{K}_a \cup \mathcal{K}''}{\!\!\!\mathbbm{1}(k, \boldsymbol{x}, \boldsymbol{z}) } =\!\!\! \prod_{k \in \mathcal{K}_a \cup \mathcal{K}''}{\!\!\!\mathbbm{1}(k, \boldsymbol{x}, \pi(\boldsymbol{x}; \boldsymbol{\theta})).} %\frac{K_a}{K}
\end{align}
In~\eqref{cell_reliability}, the cell reliability is $1$ only if all the served UEs meet their reliability target, otherwise, $ \mathbbm{1}_c(\boldsymbol{x}, \boldsymbol{\theta})=0$. Our goal is to find an optimal admission control policy that can maximize the average cell reliability, i.e., 
\begin{align}\label{opt_problem}
\max_{\boldsymbol{\theta}}	\mathbbm{E}_{\boldsymbol{x}}\left[R(\boldsymbol{x}, \boldsymbol{\theta})\right] = \max_{\boldsymbol{\theta}}	\mathbbm{E}_{\boldsymbol{x}}\left[ \frac{K_a}{K'}\mathbbm{1}_c(\boldsymbol{x}, \boldsymbol{\theta})\right],
\end{align}
where $K_a =|\mathcal{K}_a|$, $K'= |\mathcal{K}'|\neq 0$, and the expectation is with respect to random variables in $\boldsymbol{x}$. The multiplicative term  $K_a/K'$ in~\eqref{opt_problem} is included to encourage the policy to admit more applicant UEs to the cell. We can expand the objective function as: %the fading channel, noise, and random traffic.
\begin{align}\label{objective}
	\!\!\!\mathbbm{E}_{\boldsymbol{x}}\left[R(\boldsymbol{x}, \boldsymbol{\theta})\right]  \propto \int_{\boldsymbol{x}} \mathbbm{1}_c(\boldsymbol{x}, \boldsymbol{\theta}) f_{\boldsymbol{x}}(\boldsymbol{x}) d\boldsymbol{x}
	=  \int_{\boldsymbol{x} \in \mathcal{X}^{\boldsymbol{\theta}}}  f_{\boldsymbol{x}}(\boldsymbol{x}) d\boldsymbol{x},
\end{align} 
where $f_{\boldsymbol{x}}(\boldsymbol{x})$ is the joint probability density function (PDF) of variables in $\boldsymbol{x}$. Moreover, for a given set of policy parameters $\boldsymbol{\theta}$, the integral in \eqref{objective} is calculated over all $\boldsymbol{x}  \in \mathcal{X}^{\boldsymbol{\theta}}$, where the set $\mathcal{X}^{\boldsymbol{\theta}}$ is 
\begin{align}\label{set_def}
	\mathcal{X}^{\boldsymbol{\theta}} &= \left\{
	\boldsymbol{x} | \mathbbm{1}_c(\boldsymbol{x}, \boldsymbol{\theta}) \!= \!1
	\right\}\!=\!\left\{
	\boldsymbol{x} |\mathbbm{1}(k, \boldsymbol{x}, \pi(\boldsymbol{x}; \boldsymbol{\theta}))\! =\! 1, \forall k \in \mathcal{K}_a
	\right\},\notag\\  
	&\overset{\text{(a)}}{=}\left\{
	\boldsymbol{x} \bigg| \int_0^{\tau_k} f_{t_k, \boldsymbol{x}}(t, \boldsymbol{x} |\boldsymbol{\theta})dt \geq \delta f_{\boldsymbol{x}}(\boldsymbol{x}) , \forall k \in \mathcal{K}_a
	\right\}.
\end{align}
The equality (a) in~\eqref{set_def} is derived from the constraint in \eqref{ue_reliability} and by expanding $F_{t_k}(\tau_k; \boldsymbol{x}, \boldsymbol{z})$ as follows:
\begin{align*}
	F_{t_k}(\tau_k; \boldsymbol{x}, \boldsymbol{z}) &= \mathbb{P}(t_k \leq \tau_k | \boldsymbol{x}, \boldsymbol{\theta}) = \int_0^{\tau_k} \frac{f_{t_k, \boldsymbol{x}}(t, \boldsymbol{x} |\boldsymbol{\theta})}{f_{\boldsymbol{x}}(\boldsymbol{x} |\boldsymbol{\theta})}dt \\
	& = \frac{1}{f_{\boldsymbol{x}}(\boldsymbol{x})} \int_0^{\tau_k} f_{t_k, \boldsymbol{x}}(t, \boldsymbol{x} |\boldsymbol{\theta})dt.
\end{align*}
Solving the system of inequalities in \eqref{set_def} for $\boldsymbol{x}$ is difficult due to the lack of prior knowledge about the distribution functions. Therefore, standard gradient based optimization techniques cannot be applied here. To solve this problem, we will propose an ML-based solution to directly learn an effective UE admission control policy via experience, i.e., by making one-shot admission control decisions, observing the achieved cell reliability, and gradually improving the policy.

    \section{QoS-Aware UE Admission Control as a Neural Contextual Bandit Problem}\label{sec:MAB}
    To solve the proposed optimization problem, we develop a novel solution that models the admission controller xApp as a CB agent. The solution is built on the concept of  \emph{ (deep) neural CBs}~\cite{Allesiardo_NeuralCMAB}, enabling the agent to predict nonlinear rewards (to be defined as a function of the cell reliability) and accordingly, output the admission decision vector $\boldsymbol{z}$. Before we dive into the proposed solution, let's briefly overview neural CBs. 

    \subsection{Neural Contextual Bandits: Preliminaries} 
    Multi-armed bandit is a reinforcement learning framework suitable for optimizing decision-making under uncertainties. In bandit problems, an agent is presented with $A\in \mathbbm{N}^+$ choices commonly known as arms. At a given decision instance $i$, the agent selects an arm $a^i\in \mathcal{A}=\{a_0, a_1, \cdots,a_{A-1}\}$ according to some policy $\pi$ and observes a random reward $r_{a^i}\!\!\sim \!\!\nu_{a^i}$ drawn from an unknown distribution $\nu_{a^i}$. Considering $\mu_a$ as the expected reward of an arm $a \in \mathcal{A}$, the goal of the agent in bandit problems is to learn a policy that minimizes the average regret $\rho(I)$ after playing for $I$ iterations:
    \begin{align}\label{regret}
    	\rho(I) = I\mu^* - \sum_{i=1}^{I}r_{a^i},
    \end{align}
where $\mu^* = \max_{a}\mu_a$. In CBs, arm selection and the distribution of reward for each arm can depend on some additional information (context) pertaining to arms or the state of the environment. A special case of CB problems are nonlinear bandits wherein the reward is a nonlinear function of the context. While the majority of the literature has focused on linear CBs, there are a number of recent works that leverage DNNs as function approximators to capture the nonlinearity of the reward function.  That being said, a (deep) neural CB policy constitutes a DNN that maps context of each arm to a reward signal. By following an exploration strategy (e.g., epsilon greedy) and observing context-reward pairs, the agent can use gradient-based optimization to update the policy parameters $\boldsymbol{\theta}$. 

\subsection{Proposed Neural Contextual Bandit  Formulation} \label{Sec:III-B}
	%Back to the proposed problem in \eqref{opt_problem}, the objective function, representing the cell reliability, is clearly a nonlinear function of the network context $\boldsymbol{x}$. Therefore, 
	
	 To solve the problem in \eqref{opt_problem}, we can formally define the admission control as a neural CB problem as follows: 
	  
	 \emph{Arms}: in our problem, arms are defined as the subsets of applicant UEs that could be admitted by the cell. With $K'$ applicant UEs, the number of arms will be $2^{K'}$, raising scalability issues as $K'$ grows. To address this limitation, we introduce a hierarchy among UEs based on their long-term signal-to-interference-plus-noise ratio (SINR). That is, if an applicant UE $k’\in \mathcal{K}'$ is admitted by the cell, then all other applicant UEs that have a higher average SINR must also be admitted. Hence, the number of arms $A$ will be equal to $K' + 1$. As an example, consider a scenario where $K'=3$ and $\gamma_1>\gamma_3>\gamma_2$ with $\gamma_k$ denoting the average SINR for UE $k$.  Given the defined hierarchy, plausible arms will be: (a) arm $a_0$ with the decision vector $\boldsymbol{z}_{a_0}=[0,0,0]$ admitting no applicant UEs, (b) arm $a_1$ with the decision vector $\boldsymbol{z}_{a_1}=[1,0,0]$ admitting UE $k=1$ only, (c) arm $a_2$ with the decision vector $\boldsymbol{z}_{a_2 }=[1,0,1]$ admitting UEs $1$ and $3$, and (d) arm $a_3$ with the decision vector $\boldsymbol{z}_{a_3}=[1,1,1]$ admitting all three UEs.

	 \emph{Arm context}: We define a distinct feature (context) vector for each applicant UE $k'\in \mathcal{K}'$  as well as the cell.  For a UE $k'\in \mathcal{K}'$, the feature vector $\boldsymbol{x}_{k'}$ encompasses UE’s attributes such as the average effective SINR $\gamma_{k'}$ estimated from imperfect CSI feedback, packet size $b_{k'}$ in bytes, packet arrival rate $\lambda_{k'}$, and UE’s delay requirement $\tau_{k'}$. We also define a feature vector $\boldsymbol{x}_c$ for the cell, comprising the cell’s current bandwidth utilization along with features in $\boldsymbol{x}_{k''}$ averaged across existing active UEs in the set $\mathcal{K}''$. Next, the arm context for an arbitrary arm $a \in \mathcal{A}$ can be defined as 
	 \begin{align}\label{arm_context}
	 \boldsymbol{x}_{a} = \left[ \boldsymbol{x}_{k'} || \boldsymbol{x}_c \right], \forall k' \in \mathcal{K'}, \,\,\text{if} \,\, \boldsymbol{z}_{a}[k'] = 1.
	 \end{align}
 	where $[.||.]$ denotes concatenation and $\boldsymbol{z}_{a}[k'] $ is the $k'$-th element of the vector $\boldsymbol{z}_{a}$. In fact, the arm context $ \boldsymbol{x}_{a}$ is obtained by concatenating the feature vectors of all applicant UEs admitted by an arm $a$ (based on its decision vector $\boldsymbol{z}_{a}$) along with the cell’s feature vector  $\boldsymbol{x}_c$. 
	 
	  \emph{Network context}: The network context vector $\boldsymbol{x}$ can be obtained by concatenating feature vectors of all applicant UE $k'\in \mathcal{K}'$ along with the cell’s features $\boldsymbol{x}_c$, i.e., $\boldsymbol{x} = \left[ \boldsymbol{x}_{k'} || \boldsymbol{x}_c \right], \forall k' \in \mathcal{K'}$.
	
	\emph{Reward}: The reward is equal to the scaled cell reliability $R(\boldsymbol{x}, \boldsymbol{\theta})$ defined in~\eqref{opt_problem}. In real networks, the reliability indicator function per UE, defined in~\eqref{ue_reliability}, and consequently $ \mathbbm{1}_c(\boldsymbol{x}, \boldsymbol{\theta})$, are not known a priori and must be estimated via either statistical methods or empirical observations over a period of time. That is, after selecting an arm $a\in\mathcal{A}$, the network can record the performance for each admitted UE $k \in \mathcal{K}_a$ over a time window of size $T$ TTIs. Let $n_k$ be the number of packets that are transmitted to UE $k$ within this time window. Then, the network can use the following Monte Carlo (MC) approximation to estimate $ \mathbb{P}(t_k(\boldsymbol{x}_a; \boldsymbol{z}_a) \leq \tau_k)$ in \eqref{ue_reliability}:
	\begin{align}\label{approximation}
		 \mathbb{P}(t_k(\boldsymbol{x}_a; \boldsymbol{z}_a) \leq \tau_k) \approx \frac{1}{n_k} \sum_{n'=1}^{n_k} \mathbbm{1}_{t_{k,n'}(\boldsymbol{x}_a; \boldsymbol{z}_a) \leq \tau_k}.
	\end{align}
Here, $t_{k,n'}$ represents the transmission delay of $n'$-th packet for UE $k$. As $n_k$ increases, the estimation error decreases according to the well-known Dvoretzky–Kiefer–Wolfowitz (DKW) inequality. However, increasing $n_k$ leads to a higher data collection cost. To determine a suitable value for $n_k$, and accordingly the observation window size $T$, one can derive a confidence interval for $ \mathbb{P}(t_k(\boldsymbol{x}_a; \boldsymbol{z}_a) \leq \tau_k)$ using the Wilson score interval~\cite{wilson}:
\begin{align*}
	 \mathbb{P}(t_k(\boldsymbol{x}_a; \boldsymbol{z}_a) \leq \tau_k) \in \left(p_{-}, p_{+}\right) 
\end{align*}
where  
\begin{align}\label{wilson}
	p_{\pm}=\frac{n_{k,s} + 0.5\beta^2}{n_k + \beta^2}\pm\frac{\beta}{n_k + \beta^2}\sqrt{\frac{n_{k,s}n_{k,f}}{n_k} + \frac{\beta^2}{4}}.
\end{align}
In~\eqref{wilson}, $n_{k,s}$ is the number of packets successfully received within their deadline $\tau_k$, and $n_{k,f} = n_k - n_{k,s}$. Moreover, $\beta$ 
%is the appropriate value from the standard normal distribution 
is a constant that is determined based on a desired confidence level. For example, for a $99\%$ confidence interval, $\beta$ is equal to $2.58$. By considering the lower-limit $p_{-}$ of the Wilson score interval, the UE reliability can be estimated by:
\begin{align}\label{ue_reliability_estimate}
		\mathbbm{1}(k, \boldsymbol{x}, \boldsymbol{z}) \approx \begin{cases}
		1, & \text{if}\,\,  p_{-} \geq \delta,\\
		0, & \text{otherwise}.
	\end{cases} 
\end{align} 
	For evaluation purposes, the UE reliability approximation in~\eqref{ue_reliability_estimate} can be used in~\eqref{cell_reliability} to obtain the cell reliability, and accordingly, the reward $R(\boldsymbol{x}, \boldsymbol{\theta})$. 
	
	In this contextual bandit problem, the reward is a nonlinear function of the arm context $\boldsymbol{x}_{a}$. Hence, we cannot directly apply standard bandit methods, such as linear upper confidence bound (UCB), that are primarily designed for linear rewards. Next, we will introduce a solution that falls into the category of neural CBs to leverage DNNs for estimating the arms’ rewards, subject to a given context $\boldsymbol{x}_{a^i}$, $a^i \in \mathcal{A}$.
	
	 \subsection{Proposed QoS-Aware UE Admission Control}\label{sec:solution}
	To solve the proposed QoS-aware admission control problem as a neural CB, the xApp comprises a system of $K'_m$ DNNs associated with arms $\{a_1, a_2, \cdots a_{K'_m}\}$. $K'_m$ is a design parameter and can be selected based on the expected number of concurrent admission requests. Each DNN is a feedforward network comprising: (a) a linear embedding layer that transforms each input feature in $\boldsymbol{x}_a$ into a vector of length $d$, (b) a concatenation layer to stack embedded features, and (c) two additional hidden layers (with ReLU activation) followed by an output layer with the Sigmoid activation function. The embedding layer includes separate sets of trainable weights for each input feature type (packet size, delay requirement, etc.).

	Let $\boldsymbol{\theta}^{(j)}$ denote the set of parameters for DNN $j$. We note that the output size of the concatenation layer is $d$ times the length of $\boldsymbol{x}_{a_j}$ in a DNN $j$ associated with an arm $a_j$. Hence, the number of parameters varies across the DNNs.  Moreover, let $g^{(j)}(\boldsymbol{x}_{a_j})$ be the output of the $j$-th DNN. Then, the predicted cell reliability for arm $j$ will be $\hat{\mathbbm{1}}_{c, j} = \lfloor g^{(j)}(\boldsymbol{x}_{a_j}) \rceil$,  where $ \lfloor . \rceil$ denotes rounding. Accordingly, the reward per arm can be calculated as $r_{a} =\frac{K_a}{K'_m}\hat{\mathbbm{1}}_{c, j}$ for $a \in \left\{a_1, \cdots, a_{K'} \right\}$. The reason for dedicating a separate DNN to each arm is twofold. First, as mentioned above, the number of UEs admitted varies across different arms, so as the input size (length of the feature vector) of DNNs. Second, training a dedicated DNN for each arm yields a higher accuracy for cell reliability predictions.

	\begin{algorithm}[!t]
		\footnotesize
		\caption{Training of the proposed neural CB xApp}\label{Algo:1}
			\textbf{Inputs:}\,\, Number of admission control events $I$, exploration parameter $\epsilon$, learning rate $\eta$, the minimum batch size $q$, reliability requirements $\delta, \tau_k$.\\
		\textbf{Initialize}\,\, Randomly initialize parameters $\boldsymbol{\theta}^{(j)}$ and reset replay buffers $\mathcal{B}^{(a_j)}=\emptyset$ of all arms $a_j, j \in \left\{1, \cdots, K'_m\right\}$.
		\begin{algorithmic}[1]
			%\Procedure{YourProcedure}{}
			\If{an admission  event is trigerred and $i<I$}
			\State Collect  $K' \leq K'_m$ concurrent admission requests submitted to a cell along with the context $\boldsymbol{x}_a$ for arms $a \in \left\{a_1, \cdots, a_{K'} \right\}$ via the E2 interface.
			\State Estimate rewards $r_{a} =\frac{K_a}{K'_m}\hat{\mathbbm{1}}_{c, j}$ for $a \in \left\{a_1, \cdots, a_{K'} \right\}$.
			\State Generate a random value $p\sim U[0, 1]$. 
			\State If $p > \epsilon$ set $a^i =\argmax_a r_a$, otherwise, randomly select an arm.	
			\State Send the list of admitted UEs in $\mathcal{K}_{a^i}$ back to the cell for scheduling. 
			\State Observe the resulting cell reliability $\mathbbm{1}_{c,i} \triangleq\mathbbm{1}_c(\boldsymbol{x}, \boldsymbol{\theta})$ via MC over $T$ TTIs. Add the experience sample $(\boldsymbol{x}_{a^i}, \mathbbm{1}_{c,i})$ to the replay buffer $\mathcal{B}^{(a_j)}$ designated to arm $a_j = a^i$.
			\For{arms $a_j, j \in \left\{1, \cdots, K'_m\right\}$}
			 \If{$|\mathcal{B}^{(a_j)}| \geq q$}
			\State Sample a mini-batch of size $q$ samples and calculate the loss 
			\begin{align*}
				&L_{\text{BCE}}\left(\boldsymbol{\theta}^{(j)}\right) \!=\\
				&\! -\!\sum_{s=1}^q \left[ \mathbbm{1}_{c,s} \log g^{(j)}(\boldsymbol{x}_{a^s}) + (1-\mathbbm{1}_{c,s})\log\left(1-g^{(j)}(\boldsymbol{x}_{a^s})\right) \right]
			\end{align*}
			\State For each DNN, separately update the model parameters: \[\boldsymbol{\theta}^{(j)} \leftarrow \boldsymbol{\theta}^{(j)} - \eta \nabla L_{\text{BCE}}(\boldsymbol{\theta}^{(j)}). \]
			\EndIf
			\EndFor
			\EndIf
			%\EndProcedure	
		\end{algorithmic}
		\textbf{Output:}\,\,  parameters of the admission control policy $\boldsymbol{\theta}=\left\{\boldsymbol{\theta}^{(j)}\right\}_{j=1}^{K'_m}$.
	\end{algorithm}\setlength{\textfloatsep}{.5\baselineskip}
 	
	The proposed training algorithm is presented in Algorithm~\ref{Algo:1} and can be briefly described as follows.  When a new admission control event is triggered by receiving admission requests from $K'$ UEs, the arms' context  $\boldsymbol{x}_{a_j}, j \in \{1, \cdots, K'\leq K'_m\}$ are collected by the cell and sent to the admission control xApp via the E2 interface in O-RAN. Then, arms' context are fed as input to the corresponding DNNs of the xApp. During the training, the agent can observe the estimated rewards $r_a$ across arms at a decision instance $i$, and decides between choosing the best arm or randomly exploring one of the arms. The set of admitted applicants in $\mathcal{K}_{a^i}$ (according to the selected arm $a^i$) is then sent to the cell for scheduling. Next, the resulting cell reliability $\mathbbm{1}_{c,i} \triangleq\mathbbm{1}_c(\boldsymbol{x}, \boldsymbol{\theta})$ is measured over a $T$-TTI observation window. The xApp maintains separate replay buffers $\mathcal{B}^{(a_j)}$ for each arm $a_j, j \in \{1, \cdots, K'_m\}$. Thus, the experience sample $(\boldsymbol{x}_{a^i}, \mathbbm{1}_{c,i})$ is added to the replay buffer $\mathcal{B}^{(a_j)}$ designated to arm $a_j = a^i$. By comparing the estimated cell reliability of the selected arm with the observed cell reliability, a binary cross-entropy (BCE) loss function can be computed and used to calculate the gradient vectors for updating the parameters of the corresponding DNN. Specific details about mini-batch training using replay buffers are included in Algorithm~\ref{Algo:1} and learning parameters are provided in Sec.~\ref{sec:simulations}. During the inference, it can be an implementation choice to let the agent continue the training based on real-time updates or freeze the weights of DNNs. In the latter case, the agent follows a greedy approach by selecting the best performing arm in each round without doing exploration.

    \section{Simulation Results}\label{sec:simulations}

 \begin{table}[!t]%[htbp]
 	\footnotesize
 	\centering
	\caption{Simulation Parameters}
		\begin{tabular}{|d{.55}|d{.3}|}
			\hline
			%\textbf{Table}&\multicolumn{3}{|c|}{\textbf{Table Column Head}} \\
			%\cline{2-4}
		    \textbf{Parameter}& \textbf{Value} \\
			\hline
		
			Number of UEs ($K$)& $10$  \\
			\hline
			
			Number of concurrent applicant UEs ($K'$)& $\left\{1, 2, 3\right\}$  \\
			\hline
			
			Subcarrier spacing& $30$ kHz  \\
			\hline
			
			TTI duration (mini-slot)& $4$ OFDM symbols \\
			\hline
			
			Cell transmit power& $44$ dBm  \\
			\hline
			
			Allocated bandwidth to the URLLC slice & $6$ MHz  \\
			\hline
			
			Number of resource blocks (RBs) & $15$  \\
			%285kHz guardband either side
			% [6e6 - (2* 285e3) - 30e3]/30e3/12
			\hline
			
			Number of RBGs & $3$  \\
			
			\hline
			
			UE speed& $1$ m/s  \\
			\hline
			
			Packet size ($b_k$)& $\left\{0.25, \cdots, 5\right\}$ bytes  \\
			\hline
			
			Average inter-packet arrival time& $\left\{1, 2, 3\right\}$ TTIs  \\
			\hline
			
			Delay threshold ($\tau_k$)& $\left\{1, \cdots, 5\right\}$ TTIs  \\
			\hline
			
			%\multicolumn{4}{l}{$^{\mathrm{a}}$Sample of a Table footnote.}
		\end{tabular}
		\label{tab1}
\end{table}

   Simulation results are obtained using a detailed Python-based system-level simulator, supporting realistic channel models from the PyPhysim library~\cite{pyphysim}, non-ideal CSI feedback, non-full buffer traffic, link adaptation, RLC retrainsmissions, and M-LWDF scheduling. Table~\ref{tab1} lists some of the main simulation parameters related to the wireless system. To ensure that the trained ML-based admission controller generalizes to various deployment scenarios, UEs' locations are randomly initialized for each admission control instance. Moreover, traffic parameters (packet size and average arrival time) and delay requirements are randomly selected for each UE from the values reported in Table~\ref{tab1}. The xApp is implemented in PyTorch, encompassing three DNNs, as described in Sec.~\ref{sec:solution}. Each input feature is passed through a linear embedding layer of size $d=10$. The number of neurons in subsequent hidden layers are 16 and 8. 
   
   For training each DNN, agent's experiences are recorded as tuples $(\boldsymbol{x}_{a^i}, \mathbbm{1}_{c,i})$ in a dedicated replay buffer of size $500$ samples. Then, each DNN is trained using mini-batches of size $q=30$ randomly sampled from its dedicated replay buffer.  Replay buffers are continuously updated and upon reaching the capacity, the oldest experience sample is removed and a new sample is appended to the buffer. Hence, unlike supervised learning schemes, the number of training samples for each DNN will be dependent on the exploration-exploitation strategy of the agent. Here, we consider an epsilon-greedy exploration with an epsilon decay schedule $\epsilon \leftarrow \max(\epsilon * 0.99, 0.1)$ after each training step. Moreover, stochastic gradient descent (SGD) optimizer is used with a learning rate of $\eta=0.01$.
   
   To simulate admission events, UEs are randomly deployed within the coverage area of the cell. Then, applicant UEs and currently active UEs are randomly selected. To simulate the cell reliability resulting from an admission decision, we consider an MC rollout of the network for $T=30,000$ TTIs (equivalent to about $4$ seconds). From \eqref{wilson}, this observation window is large enough to estimate the reliability of $\delta=0.999$ with a $99\%$ confidence. Clearly, higher reliability targets can be assessed by increasing $T$ at the cost of more time-consuming simulations. To evaluate the regret (see Fig.~\ref{fig:convergence}), parallelization is used at an edge server to compute MC simulations concurrently for all arms. Here, the MC approach is used to obtain accurate results since as mentioned in Sec.~\ref{sec:intro}, statistical analyses do not easily extend to the generic URLLC scenarios considered in this work. Nevertheless, one could explore incorporating approximation techniques~\cite{lopez2022statistical} into the proposed solution. 
     \begin{figure}
   	\centering
   	\centerline{\includegraphics[width=\columnwidth]{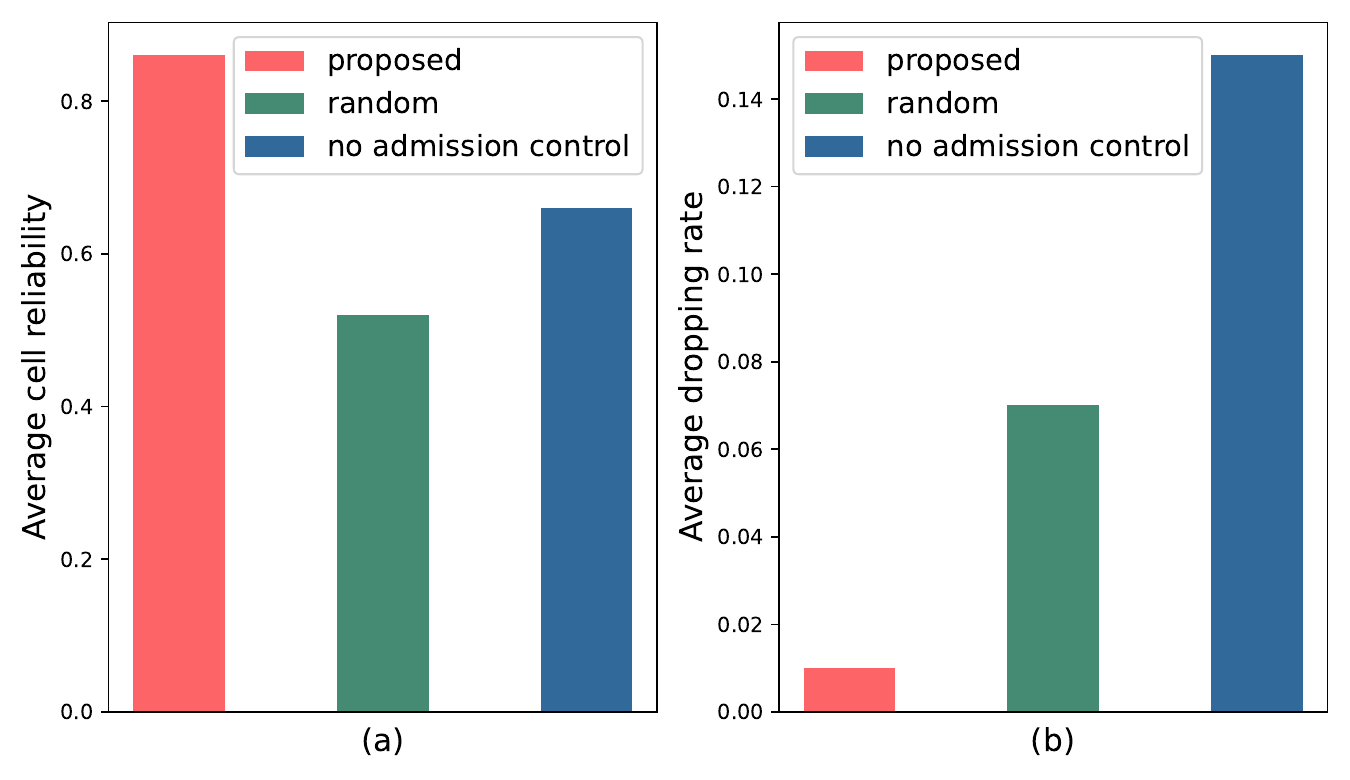}}\vspace{-1em}
   	\caption{Average cell reliability and dropping rate for different UE admission control methods.}
   	\label{fig1}
   \end{figure}

   \begin{figure}
	\centering\vspace{-.6cm}
	\centerline{\includegraphics[width=\columnwidth]{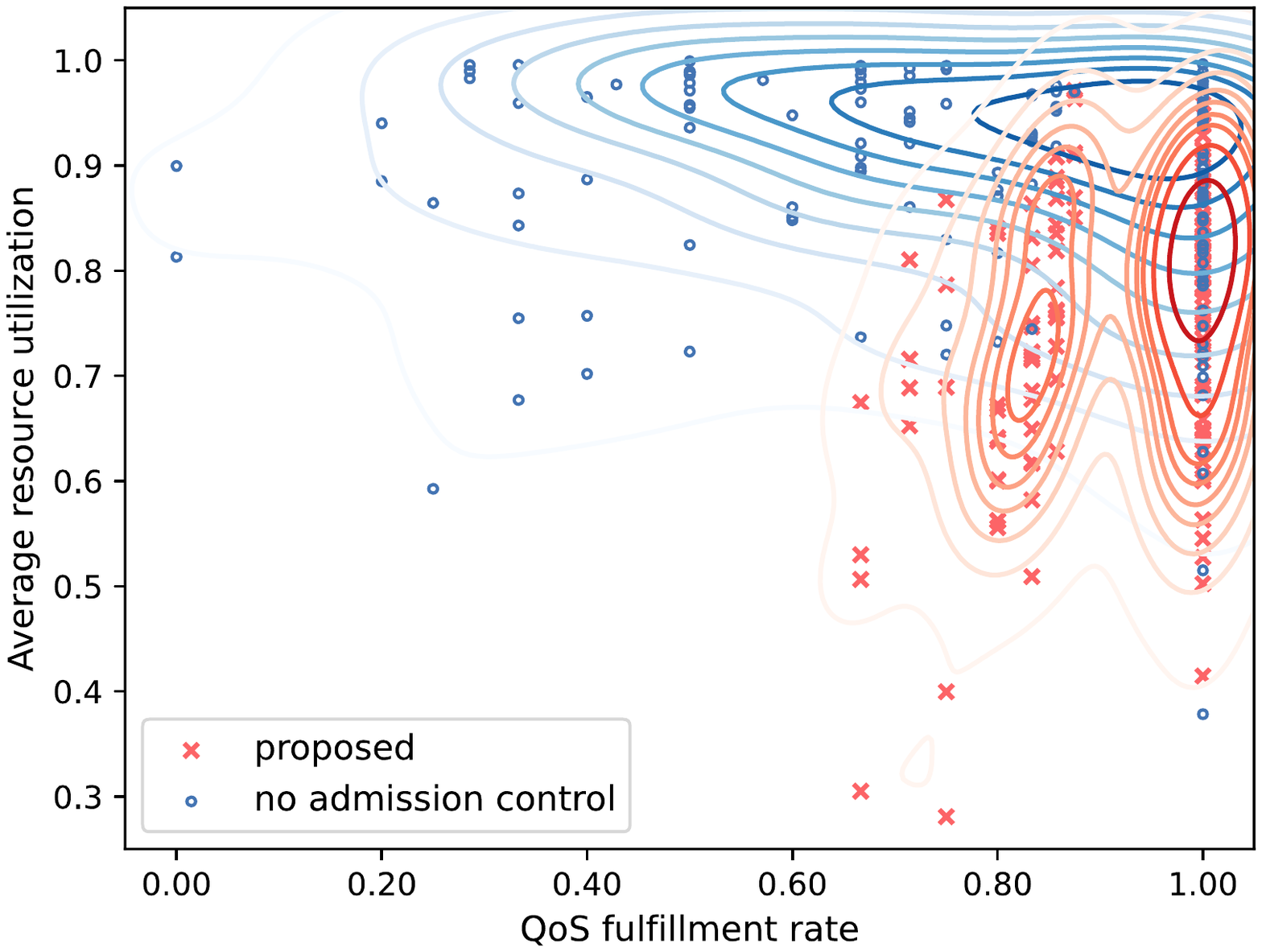}}\vspace{-1em}
	\caption{Average resource utilization vs. the QoS fulfillment rate.}
	\label{fig:sla}
\end{figure}

    For evaluations, we compare the performance of proposed approach with two other methods: (a) ``random admission control'' where an arm is randomly selected in each step, and (b) the conventional ``no admission control'' approach where no QoS-aware admission control is implemented, i.e., the cell will admit all applicant UEs.  % class imbalance and adjusted weights
   To ensure a fair performance comparison, trivial admission instances wherein the cell is already overloaded (even before admitting new UEs) are not included in our evaluations. That is because for such corner cases, the baseline methods would perform poorly. Moreover, the M-LDWF scheduling is used at the medium access control (MAC) layer for evaluating all methods. We evaluate various performance metrics, including the average cell reliability, link dropping rate, QoS fulfillment rate, and the bandwidth utilization, as discussed next.
   
   %and evaluated the performance based on various metrics. Next, we briefly describe the baselines and key performance indicators (KPIs) used for performance comparisons. Then, we present some of the simulation results demonstrating the promising aspects of the proposed scheme. We have compared the performance of the proposed approach with (a)	the ``optimal'' solution obtained by picking the best performing arm in each step, (b)	``Random admission control'' where an arm is randomly selected in each step,  (c)	``No admission control'' where any new connection request is approved by the cell. 
   
  % \subsection{Cell Reliability}
  Figure~\ref{fig1}(a) compares the cell reliability, i.e., the key performance metric defined in~\eqref{cell_reliability}, averaged over random deployments (UE locations) for different UE admission control methods. The proposed ML-based method yields $30\%$ and $65\%$ performance gains compared to the no admission control and random admission methods, respectively. The figure also shows that the proposed method achieves a near-optimal performance, i.e., for only $14\%$ of the time on average, not all admitted UEs may meet the $\delta$-reliability target.

   Figure~\ref{fig1}(b) compares the average link dropping rate for different UE admission control methods. For QoS-sensitive traffic such as in URLLC, a UE link is \emph{dropped}, i.e., terminated, if there are not sufficient available resources at the cell to meet UE's QoS~\cite{9685108}. The link dropping rate is an important performance metric in admission control and can be defined as the number of UEs that are served but their reliability target is not achieved, divided by the number of served UEs:
   \begin{align*}
   	\text{Dropping rate} =\frac{|\left\{ k \in  \mathcal{K}_{a^i} \cup 
   		\mathcal{K}_i'' | \mathbbm{1}(k, \boldsymbol{x}_{a^i}, \boldsymbol{z}_{a^i}) = 0 \right\}|}{| \mathcal{K}_{a^i} \cup 
   		\mathcal{K}_i''|}, 
   \end{align*}
    where $\mathcal{K}_i''$ is the set of active UEs at an admission decision instance $i$. Fig.~\ref{fig1}(b) shows that the proposed ML-based method completely outperforms the baselines. In fact, the proposed approach reduces the average UE dropping rate by $15$x compared to the baseline with no user admission control. The proposed method also yields a $7$x reduction in UE dropping rate compared to the baseline with random admission control. The figure also confirms the near-optimal performance of the proposed approach since only $1\%$ of the served UEs are dropped on average.   
    
   Fig.~\ref{fig:sla} shows the average resource utilization versus the average QoS fulfillment rate, for both the proposed and no admission control methods. The average resource utilization is calculated by averaging the ratio of scheduled RBGs to the total RBGs over the duration of $T$ TTIs, after making an admission control decision. The QoS fulfillment rate captures that ratio of the UEs that met their reliability target after taking an admission decision, to the maximum number of UEs that could potentially be served and satisfied, i.e.,
      \begin{align*}
   	\text{QoS ful. rate} \!=\!\frac{|\left\{ k \in \mathcal{K}_{a^i} \cup 
   		\mathcal{K}_i'' | \mathbbm{1}(k, \boldsymbol{x}_{a^i}, \boldsymbol{z}_{a^i}) = 1 \right\}|}{\max_{a^i} |\left\{ k \in  \mathcal{K}_{a^i} \!\cup\! 
   		\mathcal{K}_i'' | \mathbbm{1}(k, \boldsymbol{x}_{a^i}, \boldsymbol{z}_{a^i}) = 1 \right\}|}. 
   \end{align*}
   Note that this metric cannot be assessed by the admission control agent and only used for evaluation purpose since calculating the denominator term requires an ``oracle'' that can observe the rewards for all arms to pick the optimal arm at any admission control instance. The scatter points near bottom-right corner in Fig.~\ref{fig:sla} demonstrate the most efficient performance (high QoS fulfillment and low bandwidth utilization) while the points towards top-left represent poor admission control decisions. Further, the figure also shows the estimated distribution (solid lines) of scatter points obtained using Kernel Density Estimation (KDE) to improve visualization without adding too many scatter points. In Fig.~\ref{fig:sla}, overlapping scatter points for the proposed and no admission control schemes are removed from the evaluation, since such points mainly represent trivial ``underloaded'' scenarios wherein the optimal decision is to accept all applicant UEs, and thus, they can't contribute much to the performance comparison among the two methods. 
   
   The results in Fig.~\ref{fig:sla} show that the proposed method achieves higher QoS satisfaction while consuming less resources, thus, yielding a more efficient admission control policy. \textcolor{black}{The QoS fulfillment rate for the proposed and baseline methods are, respectively, $0.91$ and $0.76$; equivalent to $20\%$ gain in optimizing per-UE QoS satisfaction. In addition, the bandwidth utilization for the proposed and baseline methods are, respectively, $0.89$ and $0.75$; equivalent to $16\%$ bandwidth utilization reduction. Such an efficient QoS-aware UE admission control achieved by the proposed method will be imperative to meet high URLLC traffic demands in next-generation wireless networks where over-provisioning URLLC slices will no longer be a viable approach.}
   \begin{figure}
   	\centering
   	\centerline{\includegraphics[width=\columnwidth]{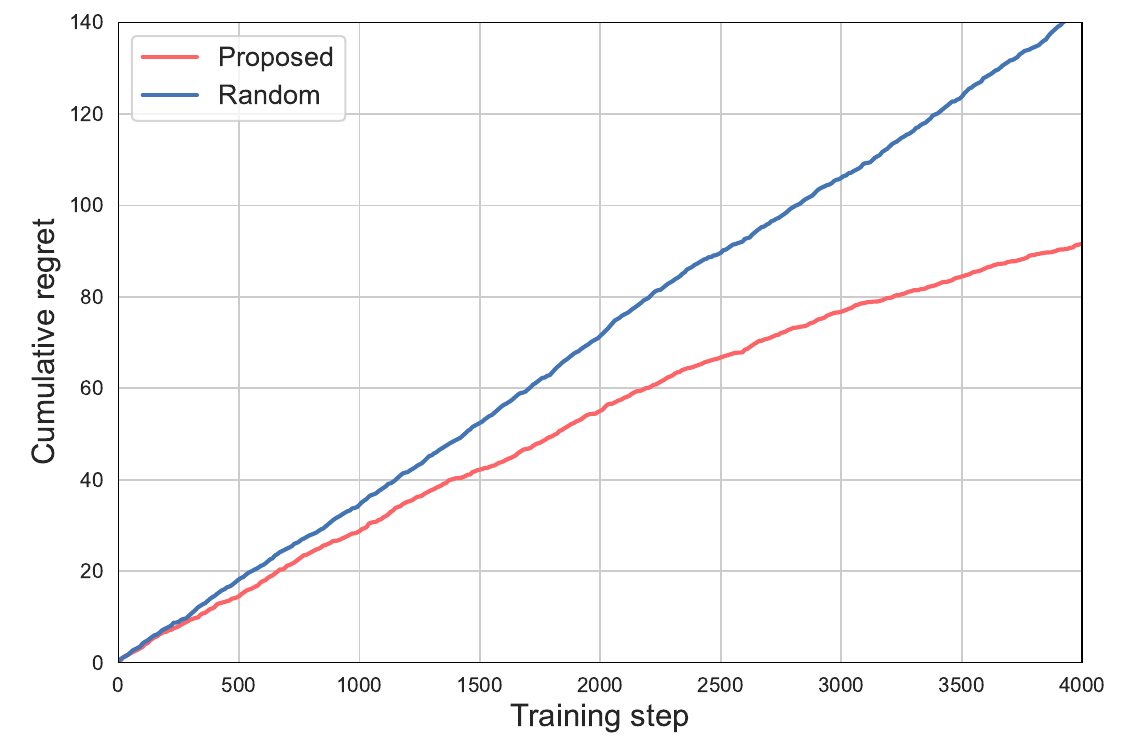}}\vspace{-1em}
   	\caption{Cumulative regret vs. training steps.}
   	\label{fig:convergence}
   \end{figure}

   Finally, Fig.~\ref{fig:convergence} compares the cumulative regret between the optimized and random bandit policies to demonstrate the effective training of the proposed neural CB admission controller. Clearly, the cumulative regret for the random method increases linearly while the curve for the proposed method is sub-linear.

    \section{Conclusions}\label{sec:conclusions}
    %\lipsum[1]
    In this paper, we have developed a novel ML-based admission control framework in O-RAN, cognizant of strict QoS requirements for the URLLC traffic. First, we have formulated a new optimization problem to maximize the average service reliability at a cell level, while considering statistical constraints associated with URLLC traffic. Then, we have solved the problem by proposing a new QoS-aware UE admission control scheme using techniques from neural CBs. The proposed solution embodies a system of DNNs, enabling the agent (xApp) to learn a complex mapping from RAN features (context) to the nonlinear cell reliability function. We have proposed an algorithm to train the agent by using the QoS predictions for each arm and following an epsilon-greedy exploration to optimize the admission control policy. We have evaluated the performance of the proposed scheme using various metrics, including the cell reliability, link dropping rate, QoS fulfillment rate, and resource utilization.  Our comprehensive simulations have demonstrated substantial performance gains for the proposed method across these metrics. \textcolor{black}{Furthermore, while we have focused on URLLC traffic, the proposed solution can be extended to other traffic types, such as guaranteed bit rate (GBR), by modifying the QoS model and RAN features.}

  %  \section*{Acknowledgment}

   % The preferred spelling citing other things!

    \bibliographystyle{IEEEtran}
    \bibliography{main.bib}

\end{document}